\documentclass{article}

\usepackage{arxiv}

\usepackage[utf8]{inputenc} 
\usepackage[T1]{fontenc}    
\usepackage{hyperref}       
\usepackage{url}            
\usepackage{booktabs}       
\usepackage{amsfonts}       
\usepackage{nicefrac}       
\usepackage{microtype}      
\usepackage{lipsum}
\usepackage{graphicx}
\usepackage{amsmath}
\graphicspath{ {./images/} }

\title{CCoE: A Compact and Efficient LLM Framework with Multi-Expert Collaboration for Resource-Limited Settings}

\author{
 Shaomang Huang \\
  Digital Security Group Qihoo 360\\
  \texttt{huangshaomang@360.cn} \\
   \And
 Jianfeng Pan \\
  Digital Security Group Qihoo 360\\
  \texttt{panjianfeng@360.cn} \\
  \And
 Min Peng \\
  Digital Security Group Qihoo 360\\
  \texttt{pengmin1@360.cn} \\
  \And
 Hanzhong Zheng \thanks{Corresponding author.} \\
  Digital Security Group Qihoo 360\\
  \texttt{zhenghanzhong@360.cn} \\
}

\begin{document}
\maketitle
\begin{abstract}
Large Language Models (LLMs) have achieved exceptional performance across diverse domains through training on massive datasets. However, scaling LLMs to support multiple downstream domain applications remains a significant challenge, especially under resource constraints. Existing approaches often struggle to balance performance across multiple domains with resource efficiency, limiting their broader applicability. To address this, we introduce the CCoE architecture, a modular framework that seamlessly integrates domain-specific experts into a unified LLM. By leveraging independently trained expert subnetworks on a shared backbone partition, CCoE achieves state-of-the-art performance while significantly reducing the resource requirements for multi-expert deployments. Furthermore, rule-based gating and expert planning in CCoE enable flexible task allocation, promoting expert collaboration to handle complex reasoning tasks. CCoE not only reduces inference costs but also provides a flexible and scalable solution for integrating domain expertise across diverse applications. Experiments on five domains demonstrate that CCoE achieves comparable performance to current domain-specific LLMs. Moreover, compared to existing multi-domain model ensemble methods, CCoE reduces memory usage by 61.3\%, while improving inference efficiency by 0.76x over parameter-efficient multi-expert integration approaches. 
\end{abstract}


\section{Introduction}
Large Language Models (LLMs) have achieved remarkable progress in a wide range of tasks \cite{chang2024survey}, driven by extensive training on high-quality datasets and alignment with human preferences \cite{christiano2017deep}. Despite these advancements, most open-source general-purpose LLMs struggle to deliver satisfactory performance in specialized domains such as code generation, mathematical reasoning, medicine, and law \cite{xu2022systematic} \cite{guha2024legalbench}. These domains typically require advanced reasoning or deep domain-specific knowledge, which general LLMs often lack due to their design focus on generality rather than specialization. To address these limitations, domain-specific supervised fine-tuning (SFT) has proven to be an effective strategy to improve LLM performance in specialized tasks, as demonstrated by recent advances in code generation \cite{rozière2024codellamaopenfoundation}, mathematical reasoning \cite{yu2024metamath}, and aligning models with domain-specific expertise \cite{Yingqiang}. However, building a model that can excel in multiple domains presents significant challenges, including resource constraints, ensuring applicability across domains, and achieving scalability for future demands.

\begin{figure}[t]
    \centering
    \includegraphics[width=0.6\columnwidth]{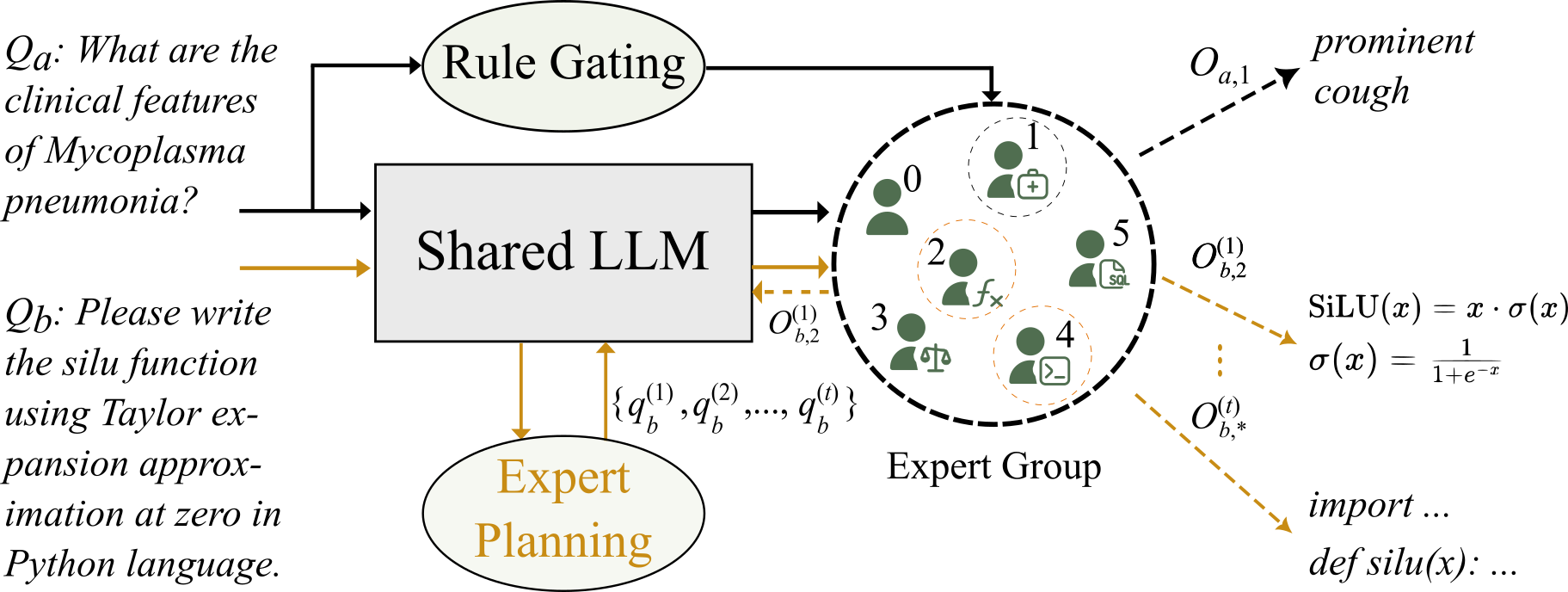}
    \caption{Examples from two typical application scenarios to illustrate the proposed CCoE framework. $Q_a$ represents a case where routing is based on a pre-identified expert (black arrow), and $O_{a,1}$ denotes the response after consulting the medical expert. $Q_b$ is a query that requires expert planning (yellow arrow), and $O^{(t)}_{b,*}$ represents the final answer obtained after collaboration among multiple experts. Note that the layers of each expert network can be flexibly distributed within the shared LLM.}
    
    \label{fig1}
    \end{figure}

    Adherence to scaling laws \cite{kaplan2020scaling}, such as increasing model size and using large and diverse multi-domain datasets, can improve the performance of general models in specialized domains \cite{lu2023emergent}. However, this approach involves high resource consumption and lacks flexibility, especially when updating specialized knowledge. Furthermore, directly extending the base model with domain-specific SFT often leads to catastrophic forgetting \cite{kotha2024understanding}, undermining the retention of prior knowledge. 
    
    Recent advances have explored the ensemble of LLMs at different scales \cite{fang2024llm} and employed agent-based collaborative systems \cite{NEURIPS2023_a3621ee9} \cite{liang2024encouraging} to develop more robust and versatile models. Using the complementary strengths of different LLMs, these methods show promise for cross-domain task reasoning. However, such approaches face critical challenges, including high memory and computational overhead, as well as deployment inefficiencies caused by inter-agent communication delays and coordination complexity. An alternative paradigm is LoRA-based \cite{hulora} efficient adapter integration, which introduces lightweight parameterized modules for domain experts while using a shared base model for joint inference \cite{hu2023llm} \cite{feng2024mixture} \cite{sheng2024slora}. Generally, these methods require the dynamic construction of a complete CUDA computation graph for each LoRA module when multiple domain experts need to be invoked, resulting in significant resource consumption and computational inefficiency.

    In this work, we propose the CCoE framework, which unifies multiple experts into a single LLM, achieving state-of-the-art performance while significantly reducing deployment costs in multi-expert applications. Each expert is implemented as a subnetwork within the backbone LLM, integrated at any specified layer, enabling it to function independently for downstream tasks. This design ensures efficiency by sharing common structures and mitigates training interference between experts. To enhance scalability, CCoE introduces a dynamic pop-and-push mechanism that enables seamless expert expansion and knowledge updates without degrading the performance of the original model. Furthermore, to accommodate diverse application scenarios, CCoE incorporates two complementary mechanisms: a rule-based gating for general real-world tasks and a planning expert for dynamic task allocation and collaboration in complex reasoning tasks. Our main contributions are as follows:
    \begin{itemize}
    \item We propose the CCoE framework, which integrates multiple domain experts into a unified architecture. By combining shared structures with independently trained expertise, it maintains efficiency while mitigating knowledge conflicts and forgetting among experts.
    \item We introduce rule-based gating and expert planning in CCoE to ensure flexibility and enhance its ability to decompose and coordinate complex tasks in real-world applications. Furthermore, CCoE can seamlessly scale the number of experts and update their knowledge through a pop-and-push training strategy. 
    \item We evaluate CCoE across five domains—Code, Math, Law, Text-to-SQL, and Medical—demonstrating performance on par with domain-specific LLMs. Furthermore, compared to multi-domain model ensemble methods, CCoE reduces memory consumption by 61.3\% and improves computational efficiency by 0.76x compared to parameter-efficient expert integration methods.
\end{itemize}

\section{Related Work}
\label{sec:rw}
\paragraph{General LLMs.} In recent years, the success of scaling laws \cite{kaplan2020scaling} has driven rapid progress in pre-trained large language models (LLMs) \cite{min2023recent}. Researchers have pushed the boundaries of performance by developing models with billions to tens of billions of parameters and training them on massive multi-domain datasets \cite{kalyan2024survey}. LLMs excel at understanding and generating natural language, matching or exceeding human-level performance on many open-domain benchmarks \cite{chang2024survey}. However, their performance in specialized domains such as medicine, law, and programming remains limited due to the domain-specific knowledge required and the scarcity of high-quality training data.

Domain-specific supervised fine-tuning (SFT) \cite{rozière2024codellamaopenfoundation} \cite{yu2024metamath} is a common strategy to address this gap but typically requires updating all model parameters, which is computationally expensive and limits flexibility. Parameter-efficient fine-tuning methods, such as LoRA \cite{hulora}, offer a more resource-efficient alternative by training only a small set of additional parameters while freezing most of the model. However, these methods \cite{hu2023llm} \cite{feng2024mixture} often suffer from catastrophic forgetting, where improvements in one domain can lead to a degradation in the model's performance in other domains. Maintaining a balance in performance across multiple domains remains a key challenge.

To address this, we propose CCoE, a compact and efficient multi-expert collaborative framework. CCoE uses a shared base model to maintain general capabilities while integrating lightweight, independently trained expert modules tailored to each domain. This design reduces parameter redundancy, minimizes knowledge interference, and improves LLM adaptability across diverse domains.

\paragraph{Multi-Domain Model Ensemble.} The multi-domain model ensemble approach improves task adaptability by combining multiple individual models, or "experts," through dynamic or static selection mechanisms that assign specific experts to domain-specific tasks. This approach addresses the limitations of a single LLM in terms of task adaptation. Researchers have explored the integration of multiple independent LLMs. For example, \cite{fang2024llm} and \cite{jiang2023llm} demonstrated that combining LLMs of different sizes can leverage their complementary strengths, resulting in improved cross-domain performance. However, these methods involve high resource costs and complex architectures, which pose challenges for practical deployment. Another line of research investigates agent-based LLM systems \cite{NEURIPS2023_a3621ee9} \cite{liang2024encouraging}, where each LLM is treated as an independent agent. These systems rely on communication protocols and collaboration mechanisms to tackle complex tasks. However, they are often constrained by inter-agent communication delays and coordination inefficiencies, resulting in high deployment costs in real-world applications.

In our work, we efficiently integrate multiple expert partitions within a single LLM, maintaining performance across multiple downstream domains while controlling model parameters. Additionally, we introduce rule-based gating for precise expert scheduling and expert planning for collaborative reasoning in complex tasks. These advances improve flexibility and applicability to address real-world challenges.

\paragraph{Efficient Multi-Expert Integration.} Efficient multi-expert integration aims to optimize computational efficiency and resource utilization while ensuring high performance through the integration of multiple experts. A notable example is the Mixture-of-Experts (MoE) framework \cite{jacobs1991adaptive} \cite{artetxe2022efficient}, which integrates multiple expert models, typically implemented as feedforward network layers. Although MoE effectively increases model capacity and reduces computational complexity during inference, it requires loading all expert parameters, resulting in substantial memory and computational overhead. Moreover, MoE faces challenges in knowledge updating and expert expansion, which limits its flexibility in dynamic scenarios.

Recently, LoRA-based parameter-efficient expert integration methods have emerged as a more economical alternative \cite{sheng2024slora} \cite{song2024multilora}. These methods can improve the performance of a base LLM in specific domains by training independent adapter modules. However, when multiple domain experts are used simultaneously, each adapter combined with the base LLM is treated as an independent computational entity, requiring the dynamic creation of a complete computation graph during inference. This redundancy significantly limits their applicability in multi-domain deployments. Furthermore, the lack of effective collaboration mechanisms between adapters limits their potential for multi-domain task applications.

In this work, we propose the CCoE framework to significantly reduce resource costs during deployment and inference, and employ a pop-and-push training strategy for flexible expert expansion and knowledge updating. Additionally, rule-based routing and expert planning enhance the flexibility and effectiveness of multi-expert applications.

\section{Method}
\subsection{Overall design}
Our CCoE framework is inspired by the concept of \textbf{C}ollaboration-\textbf{o}f-\textbf{E}xperts \cite{jacobs1991adaptive}. It consists of a shared LLM and $n$ domain-specific experts, forming a total of $n+1$ complete LLM instances. Each domain expert is implemented as a subnetwork within the framework, with its layers flexibly placed at different positions within the backbone LLM, enabling scalability of expert capacity and adaptability to diverse tasks. For ease of presentation, as shown in Figure \ref{fig1}, the expert networks are placed in the later layers of the backbone LLM in the following description.

Suppose that the backbone LLM comprises $L$ layers and incorporates $n$ domain-specific experts. The CCoE framework can be formally represented as \{$E_{b, l_b}, E_{0, l_0}, E_{1, l_1}, E_{2, l_2}, \cdots, E_{n,l_n}$\}, where $E_{b, l_b}$ denotes the shared LLM with $l_b$
layers, and $E_{i,l_i}$ denotes the $i$-th domain expert with $l_i$ layers ($i\in[0, n]$), satisfying 
 $l_b + l_0 = L$. Given an input prompt $q$, the output of the CCoE can be expressed as:
\begin{equation}
    O_i = E_{i, l_i}(E_{b, l_b}(q))\cdot\mathbf{1}_{\mathcal{R}(q)=i}
\end{equation}
Here, the shared LLM $E_{b, l_b}$ processes $q$ to generate intermediate features $E_{b, l_b}(q)$, which are then routed to the domain expert $E_{i, l_i}$ to generate the final output $O_i$. $i$ is the expert index, determined by the policy $\mathcal{R}$.

Compared to domain-specific SFT methods and MoE-based architectures, our CCoE framework supports the independent training and expansion of domain experts within a unified framework, while effectively integrating common skills via a shared LLM. This design enhances flexibility and adaptability in real-world use cases, enabling resource-efficient multi-domain expert integration.

\subsection{Routing Policy}
To dynamically determine the most suitable experts to invoke based on the input content, we define two complementary expert routing policies $\mathcal{R}$, namely rule-based gating and expert planning, to adapt to different application scenarios.

\paragraph{Rule-based Gating.} Rule-based gating offers high determinism and accuracy, making it particularly suitable for platforms that require multiple experts to be consulted simultaneously, such as in network security and intelligent education environments.

Our rule-based gating leverages a combination of the expert-domain mapping matrix $\mathcal{M}$ and the expert insertion position $\mathcal{P}$ to efficiently match tasks with experts. Specifically, given a set of domain-specific queries $Q = [q_1, q_2, \cdots, q_m]$ and a pool of candidate experts \{$E_{0, l_0}, E_{1, l_1}, E_{2, l_2}, \cdots, E_{n,l_n}$\}, the rule-based routing mechanism produces a series of execution paths $\mathrm{\tau} = [\tau_{1}, \tau_{2}, \cdots, \tau_{m}]$. Each execution path $\tau_{j}$ is represented as a tuple ($\mathcal{M}_{j,i}$, $\mathcal{P}_i$), where $\mathcal{P}_i$ denotes the layer position vector for inserting expert $i$ in the backbone network, and $\mathcal{M}_{j,i}$ is a binary vector indicating whether the query $q_j$ is associated with expert $E_{i, l_i}$, defined as:
\begin{equation}
    \mathcal{M}_{j,i}(q_j) =
    \begin{cases} 
    1, & \text{if } q_j \in \text{domain}_i, \\
    0, & \text{otherwise}.
    \end{cases}
\end{equation}
\label{equa2}

The rule-based gating process follows predefined rules that can be easily traced and interpreted. 
When a new expert is introduced to the CCoE framework, we simply add it to the mapping matrix $\mathcal{M}$ and specify its insertion position in the backbone network within $\mathcal{P}$.

\paragraph{Expert Planning.}
Expert planning dynamically selects the most suitable expert for each task, providing greater flexibility and generalization. It is particularly effective in complex scenarios that require coordination among multiple experts, such as intelligent search and cross-domain Q\&A.

For a query $q_j$ originating from an unknown domain, expert planning decomposes it into a sequence of subtasks \{${q^{(1)}_j, q^{(2)}_j, \cdots, q^{(t)}_j}$\}. Each subtask $q^{(t)}_j$ is assigned to the most suitable expert in the expert pool, with expert selection managed by a dedicated planner expert network $E_{p}$. Specifically, for each subtask, we define the matching score $h_{j,i}$ between the planner expert and each candidate expert as:
\begin{equation}
    h^{(t)}_{j,i} = \mathcal{F}(CrossAttention(W_i, E_{p}(q^{(t)}_j)))
\end{equation}
where $\mathcal{F}(\cdot)$ is a linear layer, $W_i$ is a learnable feature vector associated with the \textit{$<$indicator$>$} token for expert $i$, and $CrossAttention(\cdot)$ is similar to a self-attention layer. Here, $W_i$ serves as the query vector, while $E_{p}(q^{(t)}_j)$ provides the key and value vectors. We select the expert with the highest matching score $E_{i, l_i} = {\arg\max}_{i\in [0, n]}(h^{(t)}_{j,i})$ as the best-matched expert. The output of the subtask $q^{(t)}_j$ on this expert is then represented as:
\begin{equation}
    O^{(t)}_{j,i} = E_{i, l_i}(E_{b, l_b}([O^{(t-1)}_{j,k}; q^{(t)}_j]))
\end{equation}
where $[\cdot;\cdot]$ denotes the concatenation operation, $O^{(t-1)}_{j,k}$ represents the output of the previous subtask corresponding to expert $k$. Our expert planning uses an iterative approach, where each step's result is passed to the next expert. This enables adaptive expert selection based on specialization, improves reasoning capacity, and efficiently handles complex tasks.

\subsection{Model Training}
To enable dynamic construction and efficient updating of domain experts, the CCoE framework introduces two operations: push and pop. The push operation supports the addition or update of experts within the framework, while the pop operation facilitates the deep copying or removal of experts. Together, these operations enable flexible expert expansion and efficient memory management. Each expert’s layers are distributed across the backbone network, combining common components to optimize their training objectives, defined as:
\begin{equation}
    \mathcal{L}_{i} = - \sum^T_{t=1} \log p(y_t|y_{<t};\theta^*_b;\theta_i)
\end{equation}

Here, $y_t$ represents the target token at position $t$, $y_{<t}$ denotes all tokens generated before $t$, $\theta^*_b$ is the parameter of the shared backbone structure, which remains frozen, and $\theta_i$ refers to the parameters of the expert $i$ that need to be trained.

In practice, expert training within the CCoE framework involves two scenarios: (1) training and integrating new domain experts, and (2) continual training and optimization of existing experts. In the first scenario, a new domain-specific expert LLM ($n$) is trained on domain-specific data. Once trained, a deep copy ($n'$) is created, and its weights are transferred to the CCoE framework. The memory allocated to $n$ is then released. In the second scenario, a deep copy of the target expert ($n'$) is created for isolated training and optimized with domain-specific data. After optimization, the updated weights are transferred back to $n$, and the copy is released to free the memory. This process ensures efficient resource management and seamless integration of new knowledge.

Due to its structural flexibility, our CCoE framework can efficiently and dynamically incorporate new domains and extend existing domain knowledge without affecting the prior knowledge in the model. Compared to existing domain-specific models and integrated methods, its scalability and practicality make CCoE well-suited for integrating multiple downstream experts and enabling continuous optimization.

\section{Experiments}
We first present the five domain-specific datasets used in this study, along with the model settings and evaluation metrics. We then compare our approach to a number of state-of-the-art methods on these datasets and provide further insights through ablation studies.

\subsection{Datasets}
\label{sec:datasets}
\begin{figure}[t]
    \centering
    \includegraphics[width=0.5\columnwidth]{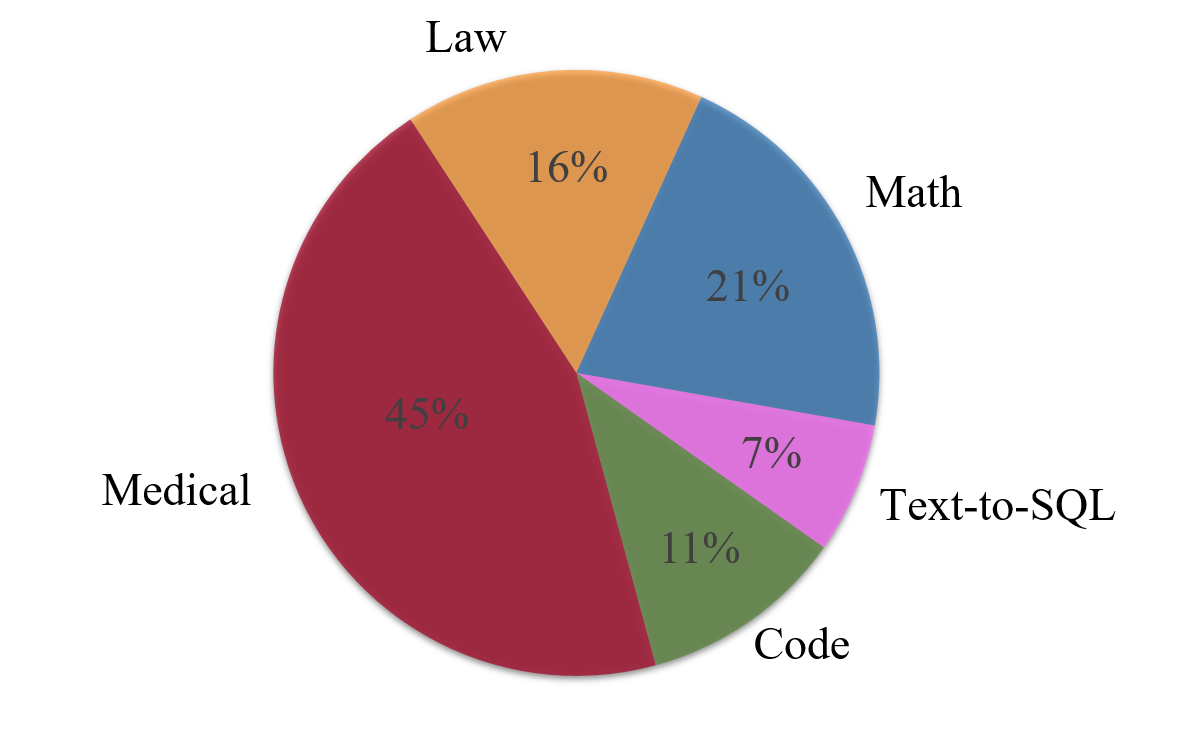}
    \caption{The distribution of data across five domains in our experimental corpus: Math, Code, Law, Medicine, and Text-to-SQL.}
    \label{fig2}
\end{figure}

We constructed experimental datasets in five domains: Math, Code, Finance, Law, Medical, and Text-to-SQL, using publicly available data. To ensure experimental reliability, we selected datasets that are widely recognized for their quality. The final distribution of these datasets is shown in Figure \ref{fig2}.

\textbf{Math.} Mathematical reasoning primarily relies on the inference capabilities of the model and is already leading the various approaches to improving performance on mathematical tasks. We selected approximately 400K instructions from the MetaMathQA \cite{yu2024metamath} dataset to train our math expert and evaluate its performance on the GSM8K\footnote{https://huggingface.co/datasets/openai/gsm8k} dataset.

\textbf{Code.} We have organized commonly used code instruction datasets, including Alpaca\footnote{https://huggingface.co/datasets/iamtarun/code\_instructions\\\_120k\_alpaca}, Code Feedback \footnote{https://huggingface.co/datasets/m-a-p/Code-Feedback}, and MBPP\footnote{https://huggingface.co/datasets/google-research-datasets/mbpp}, to construct a diverse training corpus. Our merged dataset includes several programming languages and covers important programming topics. The fine-tuning data are organized in an instruction-based format and contain more than 200K pairs. For model validation, we use MBPP test sets that do not overlap with the fine-tuning data.

\textbf{Text-to-SQL.} The Text-to-SQL (NL2SQL) dataset typically includes database tables, schemas, and QA pairs. We constructed a dataset with a total of 133K instruction-based pairs, mainly sourced from WikiSQL\footnote{https://huggingface.co/datasets/Salesforce/wikisql}, and Chase \cite{guo-etal-2021-chase}, for expert training. The model evaluation is performed on the Spider \cite{yu2018spider} dataset.

\textbf{Medical.} Our medical training dataset is mainly from MedQA \cite{jin2020disease} and HuaTuo \cite{wang2023cmb}. It consists of over 849K QA-pairs in both English and Chinese. The evaluation is performed on the MedQA test set.

\textbf{Law.} We focus on the instruction-based civil law dataset from influential projects\footnote{https://github.com/FudanDISC/DISC-LawLLM} \footnote{https://github.com/pengxiao-song/LaWGPT}. Our training dataset contains nearly 300K instruction-based pairs focused on the Chinese mainland legal system. We validate the performance of the model on the LawBench \cite{fei2023lawbench} dataset.

\subsection{Implementation Details}
\textbf{Data Preprocessing.} To efficiently preprocess the datasets, we first apply the Min-Hash \cite{broder1997resemblance} and the text-embedding-ada-002\footnote{https://openai.com/index/new-and-improved-embedding-model/} embedding model to remove highly similar questions within each domain. For the Math and Code domains, where reasoning steps are essential to improve test performance, we utilize GPT-4 to automatically generate and complete Chain-of-Thought (CoT) reasoning steps for instances lacking such annotations. For the Text-to-SQL task, we adopt the Chain-of-Hindsight (CoH) technique \cite{liu2023chainhindsightalignslanguage}, a widely adopted approach in previous studies, to construct evaluation sequences. See Appendix A.1 for more details. For all domain data, we use the Alpaca-SFT prompt template \cite{taori2023stanford} to instruct LLMs to complete the task and output the results in Markdown format.

\noindent\textbf{Experiment Settings.} 
In the CCoE framework, independently fine-tuned expert networks for each domain are integrated into a backbone LLM to address issues such as catastrophic forgetting and cross-domain interference. 

By default, we conduct our experiments using the Qwen-1.5 chat series models with 7B and 14B parameter scales as the backbone, respectively. To accommodate resource-constrained environments, the parameter size of each expert network is limited to no more than 15\% of the backbone parameters. By default, in both the 7B and 14B models, the expert layers are uniformly distributed across all decoder layer indices of the backbone network, specifically inserted at the FFN layer positions. The ablation study systematically examines alternative distribution strategies to assess their impact on model performance. We train our CCoE framework on the PyTorch platform using 8 NVIDIA H800 80G GPUs and optimize the model parameters with the AdamW optimizer. Specific hyperparameter configurations for each expert network are presented in Appendix A.2. 

\noindent\textbf{Evaluation Metrics.} We use commonly adopted metrics to evaluate model performance in different domain tasks. For Math and Medical tasks, we report accuracy (\%), where a prediction is considered correct if it matches the ground truth. For Text-to-SQL tasks, we calculate execution accuracy \cite{yu2018spider}. For Code tasks, we use Pass@k \cite{rozière2024codellamaopenfoundation}. For tasks in the Law domain, we adopt the evaluation metrics from LawBench \cite{fei2023lawbench} for various types of tasks and report their average.

\subsection{Model Evaluation}
\begin{figure}[t]
    \centering
    \includegraphics[width=0.5\columnwidth]{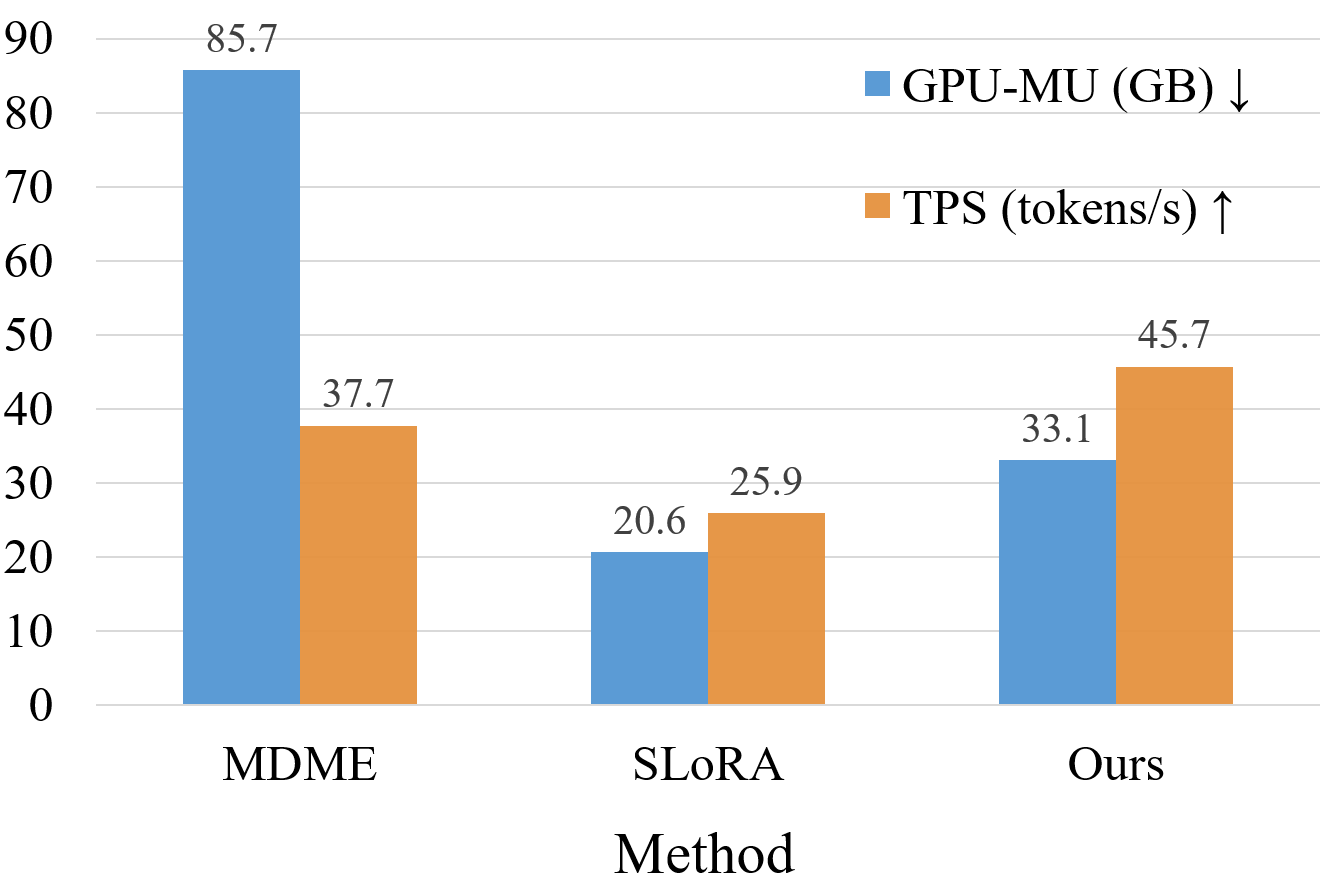}
    \caption{Comparison of GPU-MU and TPS between our CCoE framework, the MDME, and SLoRA approaches. Specifically, Our CCoE is based on a 7B backbone LLM, while MDME consists of five domain-specific models: MetaMath-7B \cite{yu2024metamath}, Fuzi-Mingcha-6B \cite{rozière2024codellamaopenfoundation}, Code LLaMA-Python-7B \cite{deng-etal-2023-syllogistic}, Meditron-7B \cite{chen2023meditron}, and RESDSQL-3B \cite{li2023resdsql}. SLoRA uses the same backbone LLM as ours and performs inference with the support of Punica LoRA in vLLM.}
    \label{fig3}
\end{figure}

\textbf{Comparison on Computational Efficiency.}
To demonstrate the resource utilization and computational efficiency advantages of our CCoE framework in multi-domain deployment scenarios, we compare its memory consumption and throughput with existing methods, including multi-domain model ensemble (MDME) \cite{fang2024llm} and parameter-efficient integration (SLoRA) \cite{sheng2024slora} in Figure \ref{fig3}. We evaluated each method by randomly sampling one question from each domain per trial. The questions were input sequentially into the model for inference, and this process was repeated 100 times to ensure statistical reliability. All evaluations were performed on a single H800 80G GPU to test resource-constrained environments. Memory consumption is measured by the maximum GPU memory usage (GPU-MU) observed during these tests, while throughput is quantified as the average number of tokens generated per second (TPS).

As shown in Figure \ref{fig3}, our CCoE framework reduces GPU-MU by approximately 61.3\% (85.7 vs 33.1) compared to the MDME method while improving inference efficiency. As the number of domain experts increases, CCoE can theoretically reduce GPU-MU by up to three-quarters compared to MDME. In contrast, MDME suffers from more frequent model switching overhead in resource-constrained scenarios, further degrading its inference efficiency.    

Furthermore, compared to SLoRA, CCoE demonstrates a significant advantage in inference efficiency, with a 0.76x increase in TPS (25.9 vs 45.7). Although SLoRA exhibits slightly lower memory usage, its need to dynamically construct computational graphs for new domain tasks in multi-domain inference scenarios considerably limits its efficiency.  

These results highlight the significant advantages of CCoE in multi-domain inference scenarios, particularly in resource-constrained environments with frequent expert switching. This further validates the practicality of the CCoE framework for multi-expert deployment, making it an efficient solution for scenarios with dynamic task requirements.

\begin{table*}[t]
\centering
\caption{The comparison of our CCoE model with existing general LLMs, domain-specific LLMs, and integration methods on Math, Code, Law, Medical, and Text-to-SQL domains.}
\resizebox{0.95\textwidth}{!}{\begin{tabular}{*8c}
\toprule
\multicolumn{4}{c}{Math} & \multicolumn{4}{c}{Law} \\
\multicolumn{2}{c}{7B model series} & \multicolumn{2}{c}{13-14B model series} & \multicolumn{2}{c}{6B-7B model series} & \multicolumn{2}{c}{13-14B model series}\\
\midrule
Baichuan2-7B & 24.5 & Baichuan2-13B & 52.8 & Baichuan2-7B & 21.2 & Baichuan2-13B & 19.1 \\
Qwen1.5-7B & 62.5 & Qwen1.5-14B & 70.1 & Qwen1.5-7B & 43.9 & Qwen1.5-14B & 49.4 \\
MathCode-L-7B & 64.2 & MathCoder-L-13B & 72.6 & Fuzi-Mingcha-6B & 28.8 & ChatLaw-13B & 32.6 \\
MetaMath-7B & 66.5 & MetaMath-13B & 72.3 & LexiLaw-6B & 26.4 & Lawyer-LLaMA-13B & 23.0 \\
Qwen1.5-7B-mix & 62.4 & Qwen1.5-14B-mix & 70.9 & Qwen1.5-7B-mix & 46.8 & Qwen1.5-14B-mix & 51.2 \\
SLoRA-7B & \underline{69.5} & SLoRA-14B & \underline{73.4} & SLoRA-7B & \underline{47.6} & SLoRA-14B & \underline{55.5} \\
\textbf{CCoE-7B Math} & \textbf{68.2} & \textbf{CCoE-14B Math} & \textbf{73.0} & \textbf{CCoE-7B Law} & \textbf{47.1} & \textbf{CCoE-14B Law} & \textbf{54.7} \\
\midrule
\midrule
\multicolumn{4}{c}{Code} & \multicolumn{2}{c}{Medical} & \multicolumn{2}{c}{Text-to-SQL}\\
\multicolumn{2}{c}{7B model series} & \multicolumn{2}{c}{13-16B model series} & \multicolumn{2}{c}{7-14B model series} & \multicolumn{2}{c}{3-14B model series}\\
\midrule
Baichuan2-7B & 24.2 & Baichuan2-13B & 30.2 & Qwen1.5-7B & 57.8 & Qwen1.5-7B & 29.9  \\
Qwen1.5-7B & 37.4 & Qwen1.5-14B & 44.0 & Qwen1.5-14B & 62.3 & Qwen1.5-14B & 34.5  \\
Code LLaMA-Python-7B & \underline{47.6} & Code LLaMA-Python-13B & \underline{49.0} & Meditron-7B & 42.8 & RESDSQL-3B+NatSQL & \underline{79.9}  \\
InCoder-6.7B & 34.4 & CodeGen-16B & 46.2 & Qwen1.5-14B-mix & 65.1 & Qwen1.5-14B-mix & 37.7 \\
Qwen1.5-7B-mix & 38.6 & Qwen1.5-14B-mix & 45.3 & SLoRA-14B & \underline{74.1}  & SLoRA-14B & 56.7 \\
SLoRA-7B & 47.2 & SLoRA-14B & 48.7 & \textbf{CCoE-7B Med} & \textbf{65.4} & \textbf{CCoE-7B SQL} & \textbf{41.8} \\
\textbf{CCoE-7B Code} & \textbf{45.5} & \textbf{CCoE-14B Code} & \textbf{48.8} & \textbf{CCoE-14B Med} & \textbf{73.5} & \textbf{CCoE-14B SQL} & \textbf{56.1}\\
\bottomrule
\end{tabular}}
\label{tab1}
\end{table*}

\noindent\textbf{Comparison with State-of-the-arts.}
Table \ref{tab1} presents a comparison of the test performance of our CCoE framework with recent open-source general LLMs (Baichuan2 \cite{yang2023baichuan}, Qwen1.5 \cite{yang2024qwen2}) and domain-specific LLMs across five domains: Math (MathCode \cite{wang2023mathcoder}, MetaMath \cite{yu2024metamath}), Code (Code LLaMA-Python \cite{rozière2024codellamaopenfoundation}, InCoder \cite{friedincoder}, CodeGen \cite{nijkampcodegen}), Law (Fuzi-Mingcha \cite{deng-etal-2023-syllogistic}, LexiLaw \cite{li2023sailer}, ChatLaw \cite{cui2023chatlaw}, Lawyer-LLaMA \cite{huang2023lawyer}), Medical (Meditron \cite{chen2023meditron}), and Text-to-SQL (RESDSQL-3B \cite{li2023resdsql}). For simplicity, our CCoE framework is evaluated under the rule-based routing setting.

Our CCoE framework consistently outperforms general LLMs in all domains, with notable improvements in Medical and Text-to-SQL tasks. These gains are primarily due to the limited coverage of these domains in the training data of general LLMs. Nevertheless, CCoE maintains a strong lead in the Math, Code, and Law domains. Compared to domain-specific LLMs, our framework outperforms most or matches the leading experts in these fields. Moreover, we compared our method with Qwen1.5-7/14B-mix (a multi-domain mixed training paradigm formed by merging training data from five domains) and parameter-efficient expert integration methods such as SLoRA. Under the same backbone parameter scale, CCoE outperforms the mixed training approach and achieves performance on par with SLoRA. We also observe that our CCoE-7B outperforms the general 14B model in most tasks.

The experiments confirm that mixed training leads to expert knowledge interference, while SLoRA benefits from global parameter tuning. These results highlight the robustness and applicability of the CCoE framework, which integrates domain-specific experts into a unified LLM, ensuring balanced and superior performance across multiple domains.

\noindent\textbf{Evaluation of Collaboration on Complex Tasks.} Here, we evaluate the performance of the CCoE framework with expert planning on complex task reasoning, which poses significant challenges for existing general-purpose LLMs, domain-specific LLMs, and integration methods. To this end, we constructed a collaborative task dataset, CoTask-12K, to train our planning expert, where each query requires the collaboration of one or more experts to complete. Table \ref{tab2} shows a typical example that requires both Math and Code expertise to solve. For more details on this dataset, please refer to Appendix B.1. We conducted comparative experiments on the GSM8K\textsubscript{Python}, MathQA\textsubscript{Numeric}, and SQL\textsubscript{Math} datasets. Successful completion of the first two tasks \cite{trung2024reft} requires knowledge of both Math and Code. The SQL\textsubscript{Math} dataset is selected from Archer \cite{zheng2024archer}, a highly challenging Text-to-SQL dataset, where each query requires the involvement of mathematical knowledge.

\begin{table}[t]\small
\centering
\caption{An example involving knowledge from the Math and Code domains.}
\resizebox{0.6\linewidth}{!}{\begin{tabular}{
l}
\hline
Question: \textit{A pet store has 15 puppies, 6 kittens, and 8 ha} \\ 
\textit{msters available for sale. Write a brief Python function} \\
\textit{snippet to determine how many different ways Alice, Bob,} \\
\textit{and Charlie can each select a different type of pet and} \\
\textit{leave the store satisfied.} \\
Labeling:\\
\textit{Participation order: $<$Math Expert$>$ $\rightarrow$ $<$Code Expert$>$} \\
  
\hline
\end{tabular}}
\label{tab2}
\end{table}

\begin{table}[t]
    \centering
     \caption{Comparative experiments on complex task collaboration.}
    \resizebox{0.7\linewidth}{!}{\begin{tabular}{*4c}
    \toprule
    Method & GSM8K\textsubscript{Python} & MathQA\textsubscript{Numeric} & SQL\textsubscript{Math}  \\
    \midrule
    Qwen1.5-7B & 27.7  & 43.1 & 10.7 \\
    Code LLaMA-Python-7B & 45.0  & 58.8 & - \\
    RESDSQL-3B & -  & - & 9.8 \\
    MDME* & 47.9  & 63.3 & 13.7 \\
    SLoRA-7B & 40.9  & 51.3 & 15.0 \\
    \textbf{CCoE-7B (Ours)} & \textbf{54.2}  & \textbf{69.2} & \textbf{17.6} \\
    \bottomrule
    \end{tabular}}
    \label{tab3}
\end{table}

The experimental results are shown in Table \ref{tab3}, with evaluation metrics consistent with those in Table \ref{tab1}. Compared to general-purpose LLM (\textit{Qwen1.5-7B}) and domain-specific models (\textit{Code LLaMA-Python-7B} and \textit{RESDSQL-3B}), our method achieves the best performance across all datasets. Moreover, our method also outperforms the multi-domain model ensemble (MDME) approach and significantly exceeds the performance of the parameter-efficient expert integration method like SLoRA. More experimental settings and qualitative results are provided in Appendix B.2.

The above experiments demonstrate that our CCoE framework enables plug-and-play integration of planning experts, effectively decomposes complex tasks, and accurately coordinates the corresponding domain experts for inference. This capability greatly improves the reasoning ability of LLMs for complex tasks. These results also highlight the CCoE framework as an effective solution for multi-domain task collaboration, further showcasing its superiority and scalability.

\begin{figure}[t]
    \centering
    \includegraphics[width=0.6\columnwidth]{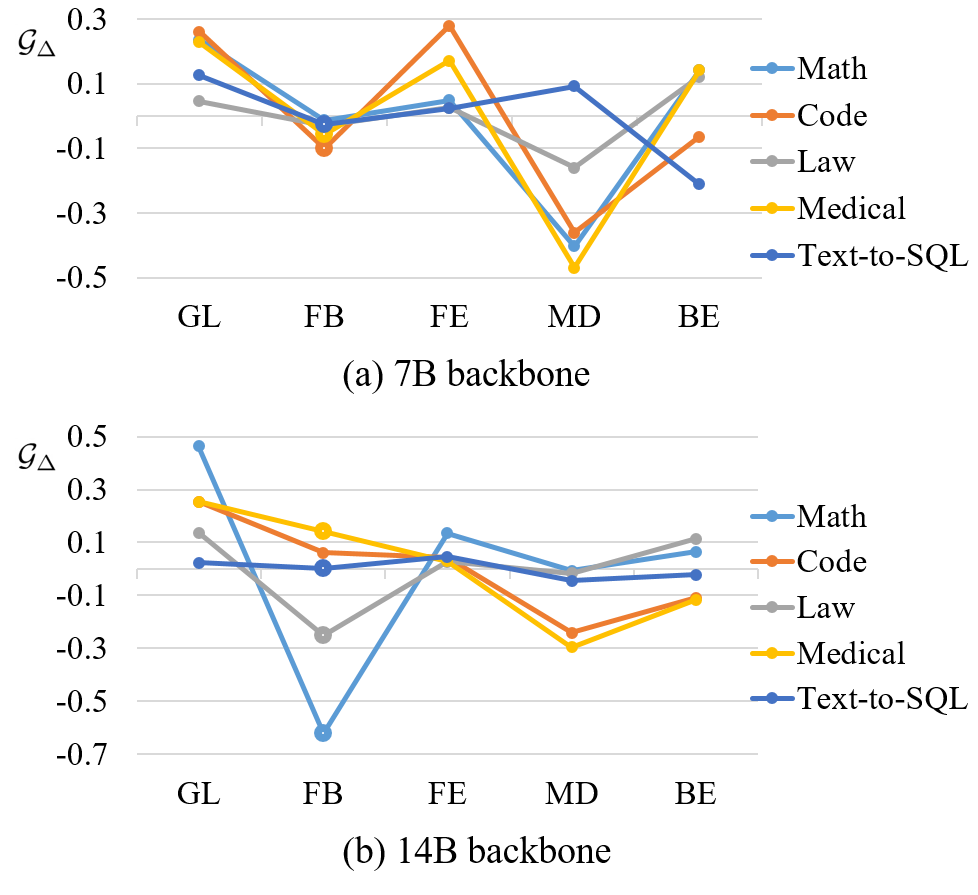}
    \caption{The evaluation of the expert layer insertion strategies in our CCoE framework.}
    \label{fig4}
\end{figure}

\subsection{Ablation Study}
\textbf{Analysis of Expert Layer Insertion Strategies.} 
In the Method section, for simplicity, we assume that the expert layers are located in the later layers of the backbone LLM. However, in practical implementations, expert layers can be distributed at different positions within the backbone, which highlights the flexibility of our CCoE framework. Here, we analyze how the distribution of expert layers affects task accuracy to explore the optimal layer distribution.

Specifically, we evaluate five distribution strategies: (1) Global (GL); (2) Front-Back (FB); (3) Front-End (FE); (4) Middle (MD); and (5) Back-End (BE), and introduce a performance gain $\mathcal{G}_\Delta$ to quantify the impact of layer distributions on expert effectiveness. Detailed explanations of these strategies and the metric $\mathcal{G}_\Delta$ are provided in Appendix C.1.

As shown in Figure \ref{fig4}, the distribution of expert layers and the metric $\mathcal{G}_\Delta$ vary by domain among different backbone models. However, in general, the best performance for all experts is achieved with the global (GL) strategy. According to studies \cite{ju2024large}, this may be because different LLM layers encode distinct linguistic patterns: shallower layers tend to capture general patterns and common sense, while deeper layers are better at encoding complex concepts. Thus, different layers of the backbone LLM play a crucial role in adapting to domain-specific knowledge.

\noindent\textbf{Comparison Between Model Families.}
In Table \ref{tab4}, we compare the performance improvements achieved by our CCoE framework when applied to different model families, relative to their respective base models. Specifically, we conducted experiments using the LLaMA2 7B chat model. The results indicate that LLaMA2 exhibits a more noticeable performance gain compared to the Qwen1.5 7B chat model, despite its overall lower performance. 

This shows that our CCoE framework can effectively adapt to various base models to enhance their domain-specific performance, while the performance ceiling is inherently influenced by the capability of the base model.

\begin{table}[t]
    \centering
     \caption{The comparison of different model families in our CCoE framework.}
    \resizebox{0.6\linewidth}{!}{\begin{tabular}{*6c}
    \toprule
    Method & CCoE-LLaMA2 7B &CCoE-Qwen1.5 7B\\
    \midrule
        Math & 36.6(+19.9)  & 68.2(+5.7)\\
        Code & 32.3(+11.5) & 45.5(+8.1)\\
        Law & 14.5(+7.35)  & 47.1(+3.3)\\
        Medical  & 30.4(+8.0)  & 65.4(+7.6)\\
        Text-to-SQL & 30.7(+16.2) &41.8(+11.9)\\
    \bottomrule
    \end{tabular}}
    \label{tab4}
\end{table}

\begin{figure}[t]
    \centering
    \includegraphics[width=0.5\columnwidth]{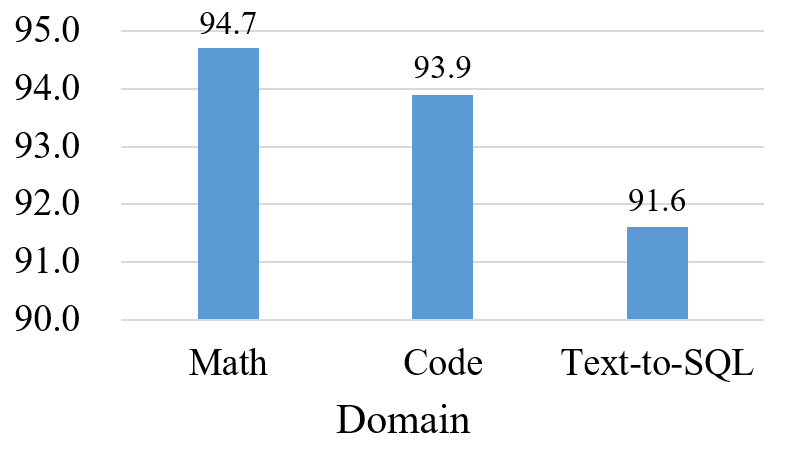}
    \caption{The evaluation of expert planning within the CCoE framework for complex tasks.}
    \label{fig5}
\end{figure}

\noindent\textbf{Analysis of expert planning capabilities.}
As shown in Table \ref{tab3}, our method performs well in reasoning complex tasks, primarily due to the expert planning in the CCoE framework, which efficiently coordinates domain experts according to task-specific requirements. To illustrate this, we randomly split the CoTask-12K dataset into training and evaluation sets in a ratio of 9:1. We believe that by scaling the training domain, the capabilities of the planning expert can be easily extended and generalized. The accuracy (\%) of our CCoE framework in complex task allocation is shown in Figure \ref{fig5}.

These results demonstrate that our CCoE framework, through expert planning, can accurately determine and invoke the required experts in the correct execution order, enabling effective collaboration between experts to perform complex task reasoning.

\section{Conclusion}
In this paper, we propose the CCoE framework, which unifies multiple expert subnetworks into a single LLM based on collaborative expertise. Extensive experiments in the Math, Code, Law, Medical, and Text-to-SQL domains show that CCoE consistently outperforms general LLMs and matches or surpasses domain-specific LLMs in evaluation. Additionally, compared to existing multi-model or parameter-efficient integration methods, CCoE reduces resource overhead and improves inference efficiency while maintaining high performance. These attributes make CCoE an effective solution for deploying multiple domain-specific LLMs in resource-constrained settings, enabling an efficient and scalable multi-expert system. Future work will explore the integration of more domain experts and build richer cross-domain datasets to further enhance the applicability and reasoning capabilities of the CCoE framework in real-world applications.

\bibliographystyle{unsrt}  

\bibliography{references}


\end{document}